\pdfoutput=1

\documentclass[11pt]{article}

\usepackage{naacl2021}

\usepackage{times}
\usepackage{latexsym}

\usepackage[T1]{fontenc}

\usepackage[utf8]{inputenc}

\usepackage{microtype}

\usepackage{soul}
\usepackage{booktabs}
\usepackage{mathtools}
\usepackage{multirow}

\newcommand{\system}[0]{TextEssence}
\newcommand{\code}[2]{#1 {\it #2}}
\newcommand{\boldcode}[2]{#1 {\bf #2}}

\title{\system{}: A Tool for Interactive Analysis of\\Semantic Shifts Between Corpora}

\author{
    Denis Newman-Griffis$^1$ \ \ \ \ \ \ \ Venkatesh Sivaraman$^2$ \ \ \ \ \ \ \ Adam Perer$^2$\\ {\bf Eric Fosler-Lussier$^3$} \ \ \ \ \ \ \ {\bf Harry Hochheiser$^1$}\\
    $^1$Department of Biomedical Informatics, University of Pittsburgh\\
    $^2$Human-Computer Interaction Institute, Carnegie Mellon University\\
    $^3$Department of Computer Science and Engineering, The Ohio State University\\
    \texttt{\{dnewmangriffis,harryh\}@pitt.edu}
}

\begin{document}
\maketitle
\begin{abstract}
Embeddings of words and concepts capture syntactic and semantic regularities of
language; however, they have seen limited use as tools to study characteristics
of different corpora and how they relate to one another. We introduce \system{},
an interactive system designed to enable comparative analysis of corpora using
embeddings. \system{} includes visual, neighbor-based, and similarity-based
modes of embedding analysis in a lightweight, web-based interface. We further
propose a new measure of embedding confidence based on nearest neighborhood
overlap, to assist in identifying high-quality embeddings for corpus analysis.
A case study on COVID-19 scientific literature illustrates the utility of
the system. \system{} is available from
\url{https://github.com/drgriffis/text-essence}.

\end{abstract}

\section{Introduction}

Distributional representations of language, such as word and concept embeddings,
provide powerful input features for NLP models in part because of their
correlation with syntactic and semantic regularities in language use
\cite{Boleda2020}. However, the use of embeddings as a lens to investigate those
regularities, and what they reveal about different text corpora, has been fairly
limited.
Prior work using embeddings to study language shifts, such as the use of
diachronic embeddings to measure semantic change in specific words over time
\cite{Hamilton2016,schlechtweg-etal-2020-semeval}, has focused primarily on
quantitative measurement of change, rather than an interactive exploration of
its qualitative aspects. On the other hand, prior work on interactive analysis
of text collections has focused on analyzing individual corpora, rather than
facilitating inter-corpus analysis \cite{liu2012tiara,weiss-2014-muck,Liu2019a}.

We introduce \system{}, a novel tool that combines the strengths of these prior
lines of research by enabling interactive comparative analysis of different text
corpora. \system{} provides a multi-view web interface for users to explore the
properties of and differences between multiple text corpora, all leveraging the
statistical correlations captured by distributional embeddings. \system{} can be
used both for categorical analysis (i.e., comparing text of different genres or
provenance) and diachronic analysis (i.e., investigating the change in a
particular type of text over time).

Our paper makes the following contributions:
\begin{itemize}
    \setlength{\itemsep}{0em}
    \item We present \system{}, a lightweight tool implemented in
    Python and the Svelte JavaScript framework, for interactive qualitative
    analysis of word and concept embeddings.
    \item We introduce a novel measure of {\it embedding confidence} to
    mitigate embedding instability and quantify the reliability of individual embedding results.
    \item We report on a case study using \system{} to investigate diachronic
    shifts in the scientific literature related to COVID-19, and demonstrate
    that \system{} captures meaningful month-to-month shifts in scientific
    discourse.
\end{itemize}

The remainder of the paper is organized as follows. \S\ref{sec:background}
lays out the conceptual background behind \system{} and its utility as a corpus
analysis tool. In \S\ref{sec:measuring-embedding-confidence} and \S\ref{sec:text-essence-system}, we describe the nearest-neighbor analysis and user interface built into \system{}. \S\ref{sec:case-study} describes our
case study on scientific literature related to COVID-19, and
\S\ref{sec:conclusion} highlights key directions for future research.

\section{Background}
\label{sec:background}

Computational analysis of text corpora can act as a lens into the social and cultural context in which those corpora were produced \cite{Nguyen2020}.
Diachronic word embeddings have been shown to reflect important context behind the corpora they are trained on, such as cultural shifts \cite{Kulkarni2015,Hamilton2016,Garg2018}, world events
\cite{kutuzov-etal-2018-diachronic}, and
changes in scientific and professional practice \cite{Vylomova2019}. However,
these analyses have proceeded independently of work on interactive tools for exploring embeddings, which are typically limited to visual projections
\cite{zhordaniya2020vec2graph,warmerdam-etal-2020-going}.
\system{} combines these directions into a single general-purpose tool
for interactively studying differences between any set of corpora, whether
categorical or diachronic.

\subsection{From words to domain concepts}
When corpora of interest are drawn from specialized domains, such as medicine, it is often necessary to shift analysis from individual words to \textit{domain concepts}, which serve to reify the shared knowledge that underpins discourse within these communities. Reified domain concepts may be referred to by multi-word surface forms (e.g., ``Lou Gehrig's disease'') and multiple distinct surface forms (e.g., ``Lou Gehrig's disease'' and ``amyotrophic lateral sclerosis''), making them more semantically powerful but also posing distinct challenges from traditional word-level representations.

A variety of embedding algorithms have been developed for learning
representations of domain concepts and real-world entities from text,
including weakly-supervised methods requiring only a terminology
\cite{Newman-Griffis2018repl4nlp}; methods using pre-trained NER models for
noisy annotation \cite{DeVine2014,Chen2020}; and methods leveraging
explicit annotations of concept mentions (as in Wikipedia)
\cite{yamada-etal-2020-wikipedia2vec}.\footnote{
    The significant literature on learning embeddings from knowledge graph
    structure is omitted here for brevity.
} These algorithms capture valuable patterns about concept types and
relationships that can inform corpus analysis \cite{runge-hovy-2020-exploring}.

\begin{figure*}
    \centering
    \includegraphics[width=0.95\textwidth]{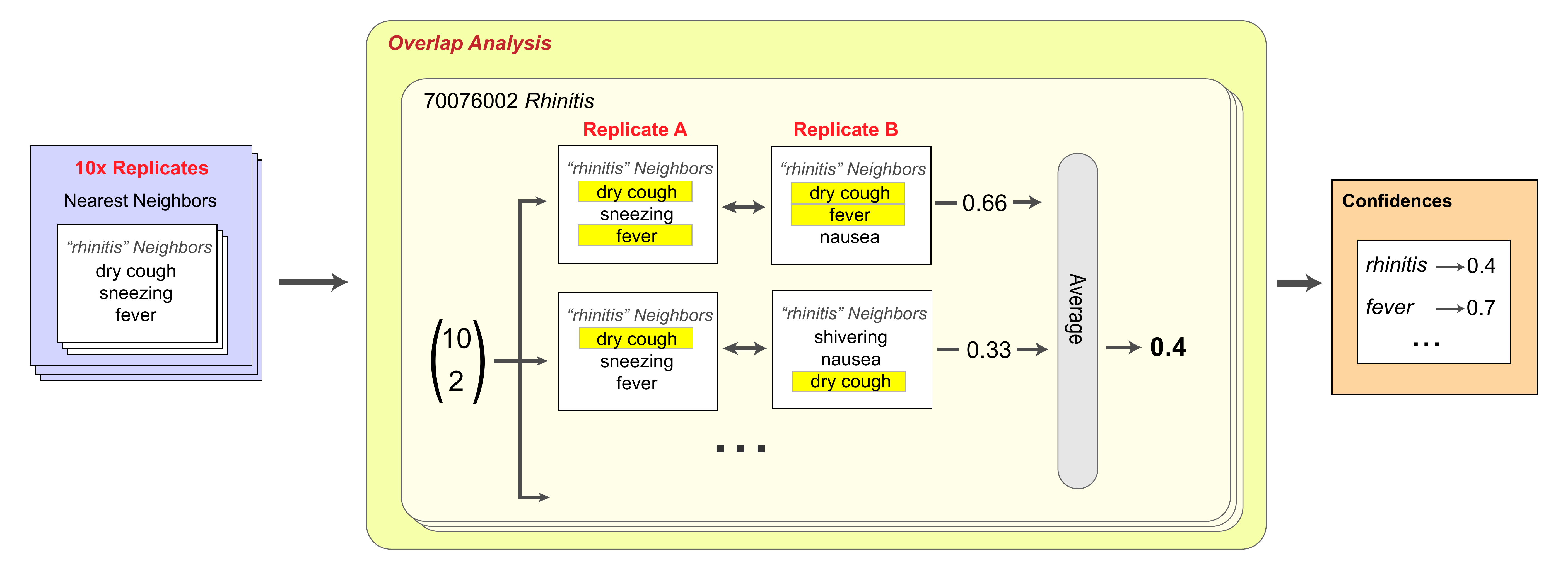}
    \caption{Illustration of embedding confidence calculation, using 10 embedding replicates.}
    \label{fig:confidence-calculation}
\end{figure*}

\system{} only requires pre-trained embeddings as input, so it can accommodate any embedding algorithm suiting the needs and characteristics of specific corpora (e.g. availability of annotations or knowledge graph resources). Furthermore, while the remainder of this paper primarily refers to concepts, \system{} can easily be used for word-level embeddings in addition to concepts.

\subsection{Why static embeddings?}
Contextualized, language model-based embeddings can provide more discriminative 
features for NLP than static (i.e., non-contextualized) embeddings. However, 
static embeddings have several advantages for this comparative use case. First, they are less resource-intensive than contextualized models, and can be efficiently trained several times without pre-training to focus entirely on the characteristics of a given corpus. Second, the scope of what static embedding methods are able to capture from a given corpus has been well-established in the literature, but is an area of current investigation for contextualized models \cite{Jawahar2019,zhao-bethard-2020-berts}. Finally, the nature of contextualized representations makes them best suited for context-sensitive tasks, while static embeddings capture aggregate patterns that lend themselves to corpus-level analysis. Nevertheless, as work on qualitative and visual analysis of contextualized models grows \cite{hoover-etal-2020-exbert}, new opportunities for comparative analysis of local contexts will provide fascinating future research.

\section{Identifying Stable Embeddings for Analysis}
\label{sec:measuring-embedding-confidence}

While embeddings are a well-established means of capturing syntax and semantics
from natural language text \cite{Boleda2020}, the problem of comparing multiple sets 
of embeddings remains an active area of research. The typical approach is to consider the nearest neighbors of specific points, consistent with the 
``similar items have similar representations'' intuition of embeddings. This method 
also avoids the conceptual difficulties and low replicability of comparing embedding 
spaces numerically (e.g. by cosine distances) \cite{gonen-etal-2020-simple}. However, 
even nearest neighborhoods are often unstable, and vary dramatically across runs of the 
same embedding algorithm on the same corpus \cite{Wendlandt2018,Antoniak2018}. In a 
setting such as our case study, the relatively small sub-corpora we use (typically 
less than 100 million tokens each) exacerbate this instability. Therefore, to quantify variation across embedding replicates and identify informative concepts, we introduce a measure of \textit{embedding confidence}.\footnote{
    An \textit{embedding replicate} here refers to the embedding matrix output
    by running a specific embedding training algorithm on a specific corpus. Ten
    runs of word2vec on a given Wikipedia dump produce ten replicates; using
    different Wikipedia dumps would produce one replicate each of ten different
    sets of embeddings.
}

We define embedding confidence as the mean overlap between the top
$k$ nearest neighbors of a given item between multiple embedding
replicates. Formally, let $E^1\dots E^m$ be $m$ embedding replicates trained on a given
corpus, and let $k\mathrm{NN}^i(c)$ be the set of $k$ nearest neighbors by cosine
similarity of concept $c$ in replicate $E^i$. Then,
the embedding confidence $EC@k$ is defined as:
    \setlength{\abovedisplayskip}{1.5pt}
    \setlength{\belowdisplayskip}{1.5pt}
\begin{multline*}
    EC@k(c,E^1\dots E^m) = \\
    \frac{1}{m(m-1)}\sum^m_i\sum_{j\neq i}
    \big|kNN^i(c)\cap kNN^j(c)\big|
\end{multline*}
This calculation is illustrated in Figure~\ref{fig:confidence-calculation}.

We can then define the set of {\it high-confidence} concepts for the given
corpus as the set of all concepts with an embedding confidence above a given
threshold. A higher threshold will restrict to highly-stable concepts only, but
exclude the majority of embeddings. We recommend an initial threshold of 0.5,
which can be configured based on observed quality of the filtered embeddings.
%

After filtering for high-confidence concepts, we summarize nearest neighbors across replicates by computing \textit{aggregate nearest neighbors}. The aggregate neighbor set of a concept $c$ is the set of high-confidence concepts with highest average cosine similarity to $c$ over the embedding replicates. This helps to provide a more reliable picture of the concept's nearest neighbors, while excluding concepts whose neighbor sets are uncertain.

\section{The \system{} Interface}
\label{sec:text-essence-system}

\begin{figure*}
    \includegraphics[width=\textwidth]{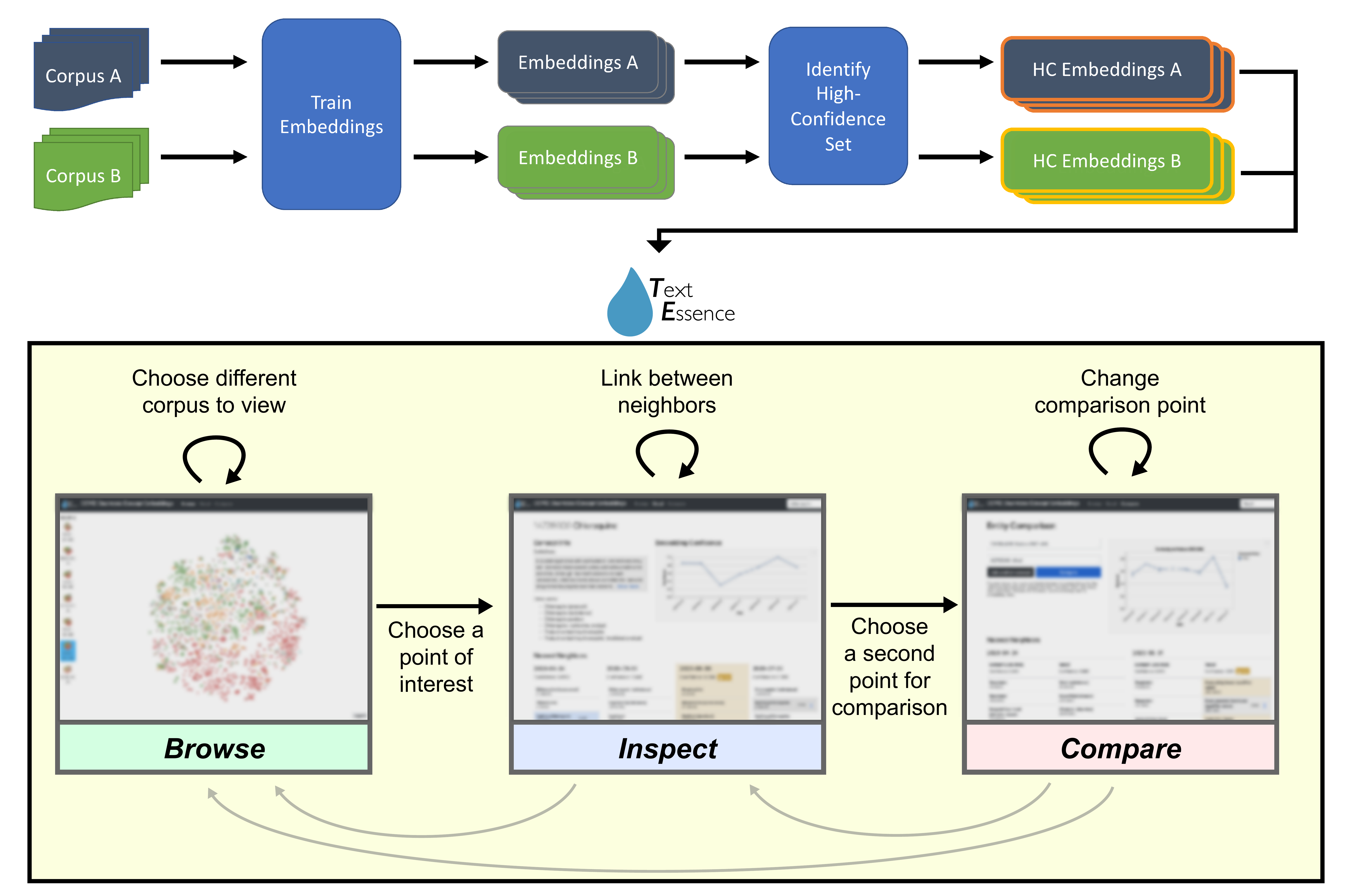}
    \caption{Workflow for comparing corpus embeddings with \system{}. The
    system enables three different kinds of interactions: (1) Browse the
    embedding space for each corpus; (2) Inspect a single concept in each
    corpus; and (3) Compare two or more concepts across corpora.
    Each view transitions to the others using the current concept.}
    \label{fig:workflow}
\end{figure*}

The workflow for using \system{} to compare different corpora is illustrated in
Figure~\ref{fig:workflow}. Given the set of corpora to compare, the user (1)
trains embedding replicates on each corpus; (2) identifies the high-confidence
set of embeddings for each corpus; and (3) provides these as input to \system{}.
\system{} then offers three modalities for interactively exploring
their learned representations: (1) \ul{Browse}, an interactive visualization of the
embedding space; (2) \ul{Inspect}, a detailed comparison of a given concept's neighbor sets across corpora; and (3) \ul{Compare}, a tool for analyzing the pairwise relationships between two or more concepts.

\subsection{Browse: visualizing embedding changes}

The first interface presented to the user is an overview visualization of one of the embedding spaces, projected into 2-D using t-distributed Stochastic Neighbor Embedding (t-SNE) \cite{tsne}. High-confidence concepts are depicted as points in a scatter plot and color-coded by their high-level semantic grouping (e.g. ``Chemicals \& Drugs,'' ``Disorders''), allowing the user to easily navigate to an area of interest. The user can select a point to highlight its aggregated nearest neighbors in the high-dimensional space, an interaction similar to TensorFlow's Embedding Projector \cite{embedding_projector} that helps distinguish true neighbors from artifacts of the dimensionality reduction process.

The Browse interface also incorporates novel interactions to address the problem of visually comparing results from several corpora (e.g., embeddings from individual months). The global structures of the corpora can differ greatly in both the high-dimensional and the low-dimensional representations, making visual comparison difficult. While previous work on comparing projected data has focused on aligning projections \cite{vis_2map,vis_comparison_word_embeddings} and adding new comparison-focused visualizations \cite{vis_compadre}, we chose to align the projections using a simple Procrustes transformation and enable the user to compare them using animation. When the user hovers on a corpus thumbnail, lines are shown between the positions of each concept in the current and destination corpora, drawing attention to the concepts that shift the most. Upon clicking the thumbnail, the points smoothly follow their trajectory lines to form the destination plot. In addition, when a concept is selected, the user can opt to \textit{center} the visualization on that point and then transition between corpora, revealing how neighboring concepts move relative to the selected one.

\subsection{Inspect: tracking individual concept change}
\label{sec:inspect-view}

Once a particular concept of interest has been identified, the Inspect view
presents an interactive table depicting how that concept's aggregated nearest neighbors have changed over time. This view also displays other contextualizing information about the concept, including its definitions (derived from the UMLS \cite{Bodenreider2004} for our case
study\footnote{
    We included definitions from all English-language sources in the UMLS, as
    SNOMED CT includes definitions only for a small subset of concepts.
}), the terms used to refer to the concept (limited to SNOMED CT for our case
study), and a visualization of the concept's embedding confidence over the sub-corpora analyzed. For information completeness, we display
nearest neighbors from every corpus
analyzed, even in corpora where the concept was not designated high-confidence
(note that a concept must be high-confidence in at least one corpus to be
selectable in the interface). In these cases, a warning is shown that the concept itself is not high-confidence in that
corpus; the neighbors themselves are still exclusively drawn from the high-confidence set.

\begin{figure}
    \includegraphics[width=0.49\textwidth]{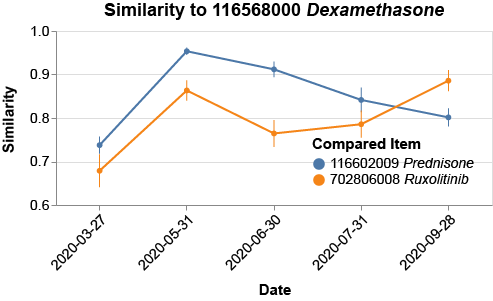}
    \caption{Similarity over time of two drugs to
    \code{116568000}{Dexamethasone} in our case study. April, August, and
    October are omitted as {\it Dexamethasone} was not high confidence for these
    months. Similarity values are mean over embedding replicates within each
    month; error bars indicate standard deviations.}
    \label{fig:pairwise-similarity}
\end{figure}

\subsection{Compare: tracking pair similarity}

The Compare view facilitates analysis of the changing relationship between two or
more concepts across corpora (e.g. from month to month).  This view displays paired nearest neighbor tables, one per corpus, showing the aggregate nearest neighbors of each of the concepts being compared. An adjacent line graph depicts the similarity between the concepts in each corpus, with one concept specified as the reference item and the others serving as comparison items (similar to Figure~\ref{fig:pairwise-similarity}). Similarity between two concepts for a specific corpus is calculated by averaging the cosine similarity between the corresponding embeddings in each replicate.

\begin{table}
    \centering
    \small
    \begin{tabular}{lcccc}
        \toprule
        Month&Docs&Words&Entities&Hi-Conf.\\
        \midrule
        March    & 41,750 &158M& 38,451 &15,100 \\
        April    & 10,738 &41M & 25,142 & 1,851 \\
        May      & 73,444 &125M& 40,297 & 5,051 \\
        June     & 24,813 &34M & 19,749 & 2,729 \\
        July     & 24,786 &35M & 19,334 & 2,800 \\
        August   & 28,642 &31M & 19,134 & 2,407 \\
        September& 33,732 &38M & 20,947 & 4,381 \\
        October  & 38,866 &44M & 21,470 & 1,990 \\
        \bottomrule
    \end{tabular}
    \caption{2020 monthly snapshots of CORD-19 dataset (documents added each month
    only; not cumulative).  Entities denotes the number of SNOMED CT codes for
    which embeddings were learned; Hi-Conf.\ is the subset of these that
    had confidence above the 0.5 threshold.
    }
    \label{tbl:CORD-19}
\end{table}

\begin{table}
    \centering
    \small
    \renewcommand{\arraystretch}{1.8}
    \begin{tabular}{p{1.7cm}p{1.2cm}p{4cm}}
        \toprule
        Concept&Month(s)&Representative neighbors\\
        \midrule
        \multirow{3}{*}{\parbox{1.7cm}{\boldcode{44169009}{Anosmia}}}
        &   Mar-Sep&
            \boldcode{2553606007}{Gustatory}, \newline
            \boldcode{51388003}{Pharyngeal pain}, \newline
            \boldcode{60707004}{Taste}\\
        &   Oct&
            \boldcode{15387003}{Vomiting}, \newline
            \boldcode{73879007}{Nausea}, \newline
            \boldcode{49727002}{Cough}\\
        \midrule
        \multirow{3}{*}{\parbox{1.7cm}{\boldcode{116568000}{Dexamethasone}}}
        &   Mar&
            \boldcode{19130008}{Injury}, \newline
            \boldcode{417746004}{Traumatic\newline injury}\\
        &   May-Jul&
            \boldcode{116602009}{Prednisone},\newline
            \boldcode{108675009}{Infliximab}\\
        &   Sep&
            \boldcode{702806008}{Ruxolitinib}\\
        \midrule
        \multirow{3}{*}{\parbox{1.7cm}{\boldcode{83490000}{Hydroxychloroquine}}}
        &   All&
            \boldcode{80229008}{Antimalarial agent},\newline
            \boldcode{96034006}{Azithromycin}\\
        &   Aug&
            \boldcode{198051006}{Nosocomial\newline infection},\newline
            \boldcode{233765002}{Respiratory failure without hypercapnia}\\
        \bottomrule
    \end{tabular}
    \caption{Representative nearest neighbors (manually selected from top 10) for three
    concepts in CORD-19, grouped by period of observation. Complete nearest
    neighbor tables are omitted for brevity, but may be viewed using our released code and data.}
    \label{tbl:findings}
\end{table}

\section{Case Study: Diachronic Change in CORD-19}
\label{sec:case-study}

The scale of global COVID-19-related research has led to an unprecedented rate
of new scientific findings, including developing understanding of the complex
relationships between drugs, symptoms, comorbidities, and health outcomes for
COVID-19 patients. We used \system{} to study how the contexts of medical
concepts in COVID-19-related scientific literature have changed over time.
Table~\ref{tbl:CORD-19} shows the number of new articles indexed in the COVID-19
Open Research Dataset (CORD-19; \citet{wang2020cord}) from its beginning in
March 2020 to the end of October 2020; while additions of new sources over time led to occasional jumps in corpus volumes, all are sufficiently large for embedding training. We created disjoint sub-corpora
containing the new articles indexed in CORD-19 each month for our case study.

CORD-19 monthly corpora were tokenized using ScispaCy \cite{Neumann2019}, and
concept embeddings were trained using JET \cite{Newman-Griffis2018repl4nlp}, a
weakly-supervised concept embedding method that does not require explicit corpus
annotations. We used SNOMED Clinical Terms (SNOMED CT), a
widely-used reference representing concepts used in clinical
reporting, as our terminology for concept embedding training,
using the March 2020 interim release of SNOMED CT International Edition, which
included COVID-19 concepts. We trained JET embeddings using a vector dimensionality
$d=100$ and 10 iterations, to reflect the relatively small size of each corpus.
We used 10 replicates per monthly corpus, and a high-confidence threshold of 0.5
for EC@5.

\subsection{Findings}

\system{} captures a number of shifts in CORD-19 that reflect how COVID-19
science has developed over the course of the pandemic.  
Table~\ref{tbl:findings} highlights key findings from our preliminary investigation into concepts known {\it a priori} to be relevant. Please note
that while full nearest neighbor tables are omitted due to space limitations,
they can be accessed by downloading our code and following the included guide to inspect CORD-19 results.

{\bf \code{44169009}{Anosmia}} While associations of anosmia (loss of sense of
smell) were observed early in the pandemic (e.g., \citet{Hornuss2020}, posted in
May 2020), it took time to begin to be utilized as a diagnostic variable
\cite{Talavera2020,Wells2020}. {\it Anosmia}'s nearest neighbors reflect this,
staying stably in the region of other otolaryngological concepts until October
(when \citet{Talavera2020,Wells2020}, {\it inter alia} were included in
CORD-19), where we observe a marked shift in utilization to play a similar role
to other common symptoms of COVID-19.

{\bf \code{116568000}{Dexamethasone}} The corticosteroid dexamethasone was
recognized early as valuable for treating severe COVID-19 symptoms
(\citet{Lester2020}, indexed July 2020), and its role has remained
stable since (\citet{Ahmed2020}, indexed October 2020). This is
reflected in the shift of its nearest neighbors from prior contexts of
traumatic brain injury \cite{Moll2020} to a stable neighborhood of
other drugs used for COVID-19 symptoms.
However, in September 2020, \code{702806008}
{Ruxolitinib} emerges as {\it Dexamethasone}'s nearest neighbor.
This reflects a spike in literature investigating the use of
ruxolitinib for severe COVID-19 symptom management \cite{Gozzetti2020,
Spadea2020,Li2020}. As the similarity graph in
Figure~\ref{fig:pairwise-similarity} shows, the contextual similarity between
dexamethasone and ruxolitinib steadily increases over time, reflecting the
growing recognition of ruxolitinib's new utility (\citet{Caocci2020}, indexed
May 2020).

{\bf \code{83490000}{Hydroxychloroquine}} Hydroxychloroquine, an anti-malarial drug, was misleadingly promoted as a potential treatment for
COVID-19 by President Trump in March, May, and July 2020, leading to
widespread misuse of the drug \cite{Englund2020}. As a result, a number of
studies re-investigated the efficacy of hydroxychloroquine as a treatment for
COVID-19\ in hospitalized patients (\citet{Ip2020,Albani2020,Rahmani2020}, all
indexed August 2020). This shift is reflected in the neighbors of
{\it Hydroxychloroquine}, adding investigative outcomes such as
nosocomial (hospital-acquired) infections and respiratory failure to the
expected anti-malarial neighbors.

\section{Conclusion and Future Work}
\label{sec:conclusion}

\system{} is an interactive tool for comparative analysis of word
and concept embeddings. Our case study on scientific literature related to
COVID-19 demonstrates that \system{} can be used to study diachronic shifts in
usage of domain concepts, and a previous study on medical records
\cite{Newman-Griffis2019louhi} showed that the technologies
behind \system{} can be used for categorical comparison as well.

The utility of \system{} is not limited to analysis of text corpora. In
settings where multiple embedding strategies are available, such as learning
representations of domain concepts from text sources \cite{Beam2020,Chen2020},
knowledge graphs \cite{Grover2016}, or both
\cite{yamada-etal-2020-wikipedia2vec,wang2020covid}, \system{} can be
used to study the different regularities captured by competing algorithms, to
gain insight into the utility of different approaches. \system{} also provides
a tool for studying the properties of different terminologies for domain
concepts, something not previously explored in the computational literature.

While our primary focus in developing \system{} was on its use as a qualitative
tool for targeted inquiry, diachronic embeddings have significant potential for
knowledge discovery through quantitative measurement of semantic differences.
However, vector-based comparison of embedding spaces faces significant
conceptual challenges, such as a lack of appropriate alignment objectives
and empirical instability \cite{gonen-etal-2020-simple}. While nearest
neighbor-based change measurement has been proposed
\cite{Newman-Griffis2019louhi,gonen-etal-2020-simple}, its
efficacy for small corpora with limited vocabularies remains to be determined.
Our novel mbedding confidence measure offers a step in this direction, but
further research is needed.

Our implementation and experimental
code is available at \url{https://github.com/drgriffis/text-essence}, and the
database derived from our CORD-19 analysis is available at \url{https://doi.org/10.5281/zenodo.4432958}. A screencast of \system{} in action is available at \url{https://youtu.be/1xEEfsMwL0k}.

\section*{Acknowledgments}
This work made use of computational resources generously provided by the Ohio
Supercomputer Center \cite{osc} in support of COVID-19 research.
The research reported in this publication was supported in part by the National
Library of Medicine of the National Institutes of Health under award number
T15 LM007059.

\bibliography{references}

\begin{thebibliography}{51}
\expandafter\ifx\csname natexlab\endcsname\relax\def\natexlab#1{#1}\fi

\bibitem[{Ahmed and Hassan(2020)}]{Ahmed2020}
Mukhtar~H Ahmed and Arez Hassan. 2020.
\newblock \href {https://doi.org/10.1007/s42399-020-00610-8} {{Dexamethasone
  for the Treatment of Coronavirus Disease (COVID-19): a Review}}.
\newblock \emph{SN Comprehensive Clinical Medicine}, 2(12):2637--2646.

\bibitem[{Albani et~al.(2020)Albani, Fusina, Giovannini, Ferretti, Granato,
  Prezioso, Divizia, Sabaini, Marri, Malpetti, and Natalini}]{Albani2020}
Filippo Albani, Federica Fusina, Alessia Giovannini, Pierluigi Ferretti, Anna
  Granato, Chiara Prezioso, Danilo Divizia, Alessandra Sabaini, Marco Marri,
  Elena Malpetti, and Giuseppe Natalini. 2020.
\newblock \href {https://doi.org/10.3390/jcm9092800} {{Impact of Azithromycin
  and/or Hydroxychloroquine on Hospital Mortality in COVID-19.}}
\newblock \emph{Journal of clinical medicine}, 9(9).

\bibitem[{Antoniak and Mimno(2018)}]{Antoniak2018}
Maria Antoniak and David Mimno. 2018.
\newblock \href {https://doi.org/10.1162/tacl_a_00008} {{Evaluating the
  Stability of Embedding-based Word Similarities}}.
\newblock \emph{Transactions of the Association for Computational Linguistics},
  6:107--119.

\bibitem[{Beam et~al.(2020)Beam, Kompa, Schmaltz, Fried, Weber, Palmer, Shi,
  Cai, and Kohane}]{Beam2020}
Andrew~L Beam, Benjamin Kompa, Allen Schmaltz, Inbar Fried, Griffin Weber,
  Nathan Palmer, Xu~Shi, Tianxi Cai, and Isaac~S Kohane. 2020.
\newblock \href {https://pubmed.ncbi.nlm.nih.gov/31797605
  https://www.ncbi.nlm.nih.gov/pmc/articles/PMC6922053/} {{Clinical Concept
  Embeddings Learned from Massive Sources of Multimodal Medical Data}}.
\newblock \emph{Pacific Symposium on Biocomputing. Pacific Symposium on
  Biocomputing}, 25:295--306.

\bibitem[{Bodenreider(2004)}]{Bodenreider2004}
Olivier Bodenreider. 2004.
\newblock \href {https://doi.org/10.1093/nar/gkh061} {{The Unified Medical
  Language System (UMLS): integrating biomedical terminology}}.
\newblock \emph{Nucleic Acids Research}, 32(90001):D267--D270.

\bibitem[{Boleda(2020)}]{Boleda2020}
Gemma Boleda. 2020.
\newblock \href {https://doi.org/10.1146/annurev-linguistics-011619-030303}
  {{Distributional Semantics and Linguistic Theory}}.
\newblock \emph{Annual Review of Linguistics}, 6(1):213--234.

\bibitem[{Caocci and {La Nasa}(2020)}]{Caocci2020}
Giovanni Caocci and Giorgio {La Nasa}. 2020.
\newblock \href {https://doi.org/10.1007/s00277-020-04067-6} {{Could
  ruxolitinib be effective in patients with COVID-19 infection at risk of acute
  respiratory distress syndrome (ARDS)?}}

\bibitem[{Chen et~al.(2018)Chen, Tao, and Lin}]{vis_comparison_word_embeddings}
Juntian Chen, Yubo Tao, and Hai Lin. 2018.
\newblock \href {https://doi.org/https://doi.org/10.1016/j.jvlc.2018.08.008}
  {Visual exploration and comparison of word embeddings}.
\newblock \emph{Journal of Visual Languages \& Computing}, 48:178 -- 186.

\bibitem[{Chen et~al.(2020)Chen, Lee, Yan, Kim, Wei, and Lu}]{Chen2020}
Qingyu Chen, Kyubum Lee, Shankai Yan, Sun Kim, Chih-Hsuan Wei, and Zhiyong Lu.
  2020.
\newblock \href {https://doi.org/10.1371/journal.pcbi.1007617} {{BioConceptVec:
  Creating and evaluating literature-based biomedical concept embeddings on a
  large scale}}.
\newblock \emph{PLOS Computational Biology}, 16(4):e1007617.

\bibitem[{Cutura et~al.(2020)Cutura, Aupetit, Fekete, and
  Sedlmair}]{vis_compadre}
Rene Cutura, Micha\"{e}l Aupetit, Jean-Daniel Fekete, and Michael Sedlmair.
  2020.
\newblock \href {https://doi.org/10.1145/3399715.3399875} {Comparing and
  exploring high-dimensional data with dimensionality reduction algorithms and
  matrix visualizations}.
\newblock In \emph{Proceedings of the International Conference on Advanced
  Visual Interfaces}, AVI '20, New York, NY, USA. Association for Computing
  Machinery.

\bibitem[{{De Vine} et~al.(2014){De Vine}, Zuccon, Koopman, Sitbon, and
  Bruza}]{DeVine2014}
Lance {De Vine}, Guido Zuccon, Bevan Koopman, Laurianne Sitbon, and Peter
  Bruza. 2014.
\newblock \href {https://doi.org/10.1145/2661829.2661974} {{Medical semantic
  similarity with a neural language model}}.
\newblock In \emph{Proceedings of the 23rd ACM International Conference on
  Information and Knowledge Management - CIKM '14}, CIKM '14, pages 1819--1822,
  Shanghai, China. ACM.

\bibitem[{Englund et~al.(2020)Englund, Kinlaw, and Sheikh}]{Englund2020}
Tessa~R Englund, Alan~C Kinlaw, and Saira~Z Sheikh. 2020.
\newblock \href {https://doi.org/https://doi.org/10.1002/acr2.11207} {{Rise and
  Fall: Hydroxychloroquine and COVID-19 Global Trends: Interest, Political
  Influence, and Potential Implications}}.
\newblock \emph{ACR Open Rheumatology}, 2(12):760--766.

\bibitem[{Garg et~al.(2018)Garg, Schiebinger, Jurafsky, and Zou}]{Garg2018}
Nikhil Garg, Londa Schiebinger, Dan Jurafsky, and James Zou. 2018.
\newblock \href {https://doi.org/10.1073/pnas.1720347115} {{Word embeddings
  quantify 100 years of gender and ethnic stereotypes}}.
\newblock \emph{Proceedings of the National Academy of Sciences},
  115(16):E3635----E3644.

\bibitem[{Gonen et~al.(2020)Gonen, Jawahar, Seddah, and
  Goldberg}]{gonen-etal-2020-simple}
Hila Gonen, Ganesh Jawahar, Djam{\'{e}} Seddah, and Yoav Goldberg. 2020.
\newblock \href {https://doi.org/10.18653/v1/2020.acl-main.51} {{Simple,
  Interpretable and Stable Method for Detecting Words with Usage Change across
  Corpora}}.
\newblock In \emph{Proceedings of the 58th Annual Meeting of the Association
  for Computational Linguistics}, pages 538--555, Online. Association for
  Computational Linguistics.

\bibitem[{Gozzetti et~al.(2020)Gozzetti, Capochiani, and
  Bocchia}]{Gozzetti2020}
Alessandro Gozzetti, Enrico Capochiani, and Monica Bocchia. 2020.
\newblock \href {https://doi.org/10.1038/s41375-020-01038-8} {{The Janus kinase
  1/2 inhibitor ruxolitinib in COVID-19.}}

\bibitem[{Grover and Leskovec(2016)}]{Grover2016}
Aditya Grover and Jure Leskovec. 2016.
\newblock \href {https://doi.org/10.1145/2939672.2939754} {{Node2Vec: Scalable
  Feature Learning for Networks}}.
\newblock In \emph{Proceedings of the 22Nd ACM SIGKDD International Conference
  on Knowledge Discovery and Data Mining}, KDD '16, pages 855--864, New York,
  NY, USA. ACM.

\bibitem[{Hamilton et~al.(2016)Hamilton, Leskovec, and Jurafsky}]{Hamilton2016}
William~L. Hamilton, Jure Leskovec, and Dan Jurafsky. 2016.
\newblock \href {http://arxiv.org/abs/1605.09096} {{Diachronic Word Embeddings
  Reveal Statistical Laws of Semantic Change}}.
\newblock In \emph{Proceedings of the 54th Annual Meeting of the Association
  for Computational Linguistics (Volume 1: Long Papers)}, pages 1489--1501,
  Berlin, Germany. Association for Computational Linguistics.

\bibitem[{Hoover et~al.(2020)Hoover, Strobelt, and
  Gehrmann}]{hoover-etal-2020-exbert}
Benjamin Hoover, Hendrik Strobelt, and Sebastian Gehrmann. 2020.
\newblock \href {https://doi.org/10.18653/v1/2020.acl-demos.22}
  {{ex{\{}BERT{\}}: {\{}A{\}} {\{}V{\}}isual {\{}A{\}}nalysis {\{}T{\}}ool to
  {\{}E{\}}xplore {\{}L{\}}earned {\{}R{\}}epresentations in
  {\{}T{\}}ransformer {\{}M{\}}odels}}.
\newblock In \emph{Proceedings of the 58th Annual Meeting of the Association
  for Computational Linguistics: System Demonstrations}, pages 187--196,
  Online. Association for Computational Linguistics.

\bibitem[{Hornuss et~al.(2020)Hornuss, Lange, Schr{\"{o}}ter, Rieg, Kern, and
  Wagner}]{Hornuss2020}
Daniel Hornuss, Berit Lange, Nils Schr{\"{o}}ter, Siegbert Rieg, Winfried~V
  Kern, and Dirk Wagner. 2020.
\newblock \href {https://doi.org/10.1101/2020.04.28.20083311} {{Anosmia in
  COVID-19 patients}}.
\newblock \emph{medRxiv}, page 2020.04.28.20083311.

\bibitem[{Ip et~al.(2020)Ip, Berry, Hansen, Goy, Pecora, Sinclaire, Bednarz,
  Marafelias, Berry, Berry, Mathura, Sawczuk, Biran, Go, Sperber, Piwoz,
  Balani, Cicogna, Sebti, Zuckerman, Rose, Tank, Jacobs, Korcak, Timmapuri,
  Underwood, Sugalski, Barsky, Varga, Asif, Landolfi, and Goldberg}]{Ip2020}
Andrew Ip, Donald~A Berry, Eric Hansen, Andre~H Goy, Andrew~L Pecora,
  Brittany~A Sinclaire, Urszula Bednarz, Michael Marafelias, Scott~M Berry,
  Nicholas~S Berry, Shivam Mathura, Ihor~S Sawczuk, Noa Biran, Ronaldo~C Go,
  Steven Sperber, Julia~A Piwoz, Bindu Balani, Cristina Cicogna, Rani Sebti,
  Jerry Zuckerman, Keith~M Rose, Lisa Tank, Laurie~G Jacobs, Jason Korcak,
  Sarah~L Timmapuri, Joseph~P Underwood, Gregory Sugalski, Carol Barsky,
  Daniel~W Varga, Arif Asif, Joseph~C Landolfi, and Stuart~L Goldberg. 2020.
\newblock \href {https://doi.org/10.1371/journal.pone.0237693}
  {{Hydroxychloroquine and tocilizumab therapy in COVID-19 patients-An
  observational study.}}
\newblock \emph{PloS one}, 15(8):e0237693.

\bibitem[{Jawahar et~al.(2019)Jawahar, Sagot, and Seddah}]{Jawahar2019}
Ganesh Jawahar, Beno$\backslash${\^{}}$\backslash$it Sagot, and Djam{\'{e}}
  Seddah. 2019.
\newblock \href {https://doi.org/10.18653/v1/P19-1356} {{What Does {\{}BERT{\}}
  Learn about the Structure of Language?}}
\newblock In \emph{Proceedings of the 57th Annual Meeting of the Association
  for Computational Linguistics}, pages 3651--3657, Florence, Italy.
  Association for Computational Linguistics.

\bibitem[{Kulkarni et~al.(2015)Kulkarni, Al-Rfou, Perozzi, and
  Skiena}]{Kulkarni2015}
Vivek Kulkarni, Rami Al-Rfou, Bryan Perozzi, and Steven Skiena. 2015.
\newblock \href {https://doi.org/10.1145/2736277.2741627} {{Statistically
  Significant Detection of Linguistic Change}}.
\newblock In \emph{Proceedings of the 24th International Conference on World
  Wide Web}, WWW '15, pages 625--635, Republic and Canton of Geneva, CHE.
  International World Wide Web Conferences Steering Committee.

\bibitem[{Kutuzov et~al.(2018)Kutuzov, {\O}vrelid, Szymanski, and
  Velldal}]{kutuzov-etal-2018-diachronic}
Andrey Kutuzov, Lilja {\O}vrelid, Terrence Szymanski, and Erik Velldal. 2018.
\newblock \href {https://www.aclweb.org/anthology/C18-1117} {{Diachronic word
  embeddings and semantic shifts: a survey}}.
\newblock In \emph{Proceedings of the 27th International Conference on
  Computational Linguistics}, pages 1384--1397, Santa Fe, New Mexico, USA.
  Association for Computational Linguistics.

\bibitem[{Lester et~al.(2020)Lester, Sahin, and Pasyar}]{Lester2020}
Mohammed Lester, Ali Sahin, and Ali Pasyar. 2020.
\newblock \href {https://doi.org/https://doi.org/10.1016/j.amsu.2020.07.004}
  {{The use of dexamethasone in the treatment of COVID-19}}.
\newblock \emph{Annals of Medicine and Surgery}, 56:218--219.

\bibitem[{Li and Liu(2020)}]{Li2020}
Hailan Li and Huaping Liu. 2020.
\newblock \href {https://doi.org/10.1016/j.jaci.2020.09.002} {{Whether the
  timing of patient randomization interferes with the assessment of the
  efficacy of ruxolitinib for severe COVID-19}}.
\newblock \emph{Journal of Allergy and Clinical Immunology}, 146(6):1453.

\bibitem[{Liu et~al.(2019)Liu, Wang, Collins, Dou, Ouyang, El-Assady, Jiang,
  and Keim}]{Liu2019a}
S~Liu, X~Wang, C~Collins, W~Dou, F~Ouyang, M~El-Assady, L~Jiang, and D~A Keim.
  2019.
\newblock \href {https://doi.org/10.1109/TVCG.2018.2834341} {{Bridging Text
  Visualization and Mining: A Task-Driven Survey}}.
\newblock \emph{IEEE Transactions on Visualization and Computer Graphics},
  25(7):2482--2504.

\bibitem[{Liu et~al.(2012)Liu, Zhou, Pan, Song, Qian, Cai, and
  Lian}]{liu2012tiara}
Shixia Liu, Michelle~X Zhou, Shimei Pan, Yangqiu Song, Weihong Qian, Weijia
  Cai, and Xiaoxiao Lian. 2012.
\newblock {Tiara: Interactive, topic-based visual text summarization and
  analysis}.
\newblock \emph{ACM Transactions on Intelligent Systems and Technology (TIST)},
  3(2):1--28.

\bibitem[{{Liu} et~al.(2020){Liu}, {Zhang}, {Leontie}, {Stylianou}, and
  {Pless}}]{vis_2map}
X.~{Liu}, Z.~{Zhang}, R.~{Leontie}, A.~{Stylianou}, and R.~{Pless}. 2020.
\newblock \href {https://doi.org/10.1109/WACV45572.2020.9093469} {2-map:
  Aligned visualizations for comparison of high-dimensional point sets}.
\newblock In \emph{2020 IEEE Winter Conference on Applications of Computer
  Vision (WACV)}, pages 2539--2547.

\bibitem[{Moll et~al.(2020)Moll, Lara, Pomar, Orozco, Frontera, Llompart-Pou,
  Moratinos, Gonz{\'{a}}lez, Ib{\'{a}}{\~{n}}ez, and
  P{\'{e}}rez-B{\'{a}}rcena}]{Moll2020}
Apolonia Moll, M{\'{o}}nica Lara, Jaume Pomar, M{\'{o}}nica Orozco, Guiem
  Frontera, Juan~A Llompart-Pou, Lesmes Moratinos, V{\'{i}}ctor Gonz{\'{a}}lez,
  Javier Ib{\'{a}}{\~{n}}ez, and Jon P{\'{e}}rez-B{\'{a}}rcena. 2020.
\newblock \href
  {https://journals.lww.com/md-journal/Fulltext/2020/10230/Effects{\_}of{\_}dexamethasone{\_}in{\_}traumatic{\_}brain{\_}injury.106.aspx}
  {{Effects of dexamethasone in traumatic brain injury patients with
  pericontusional vasogenic edema: A prospective-observational DTI-MRI study}}.
\newblock \emph{Medicine}, 99(43).

\bibitem[{Neumann et~al.(2019)Neumann, King, Beltagy, and Ammar}]{Neumann2019}
Mark Neumann, Daniel King, Iz~Beltagy, and Waleed Ammar. 2019.
\newblock \href {https://doi.org/10.18653/v1/W19-5034}
  {{{\{}S{\}}cispa{\{}C{\}}y: Fast and Robust Models for Biomedical Natural
  Language Processing}}.
\newblock In \emph{Proceedings of the 18th BioNLP Workshop and Shared Task},
  pages 319--327, Florence, Italy. Association for Computational Linguistics.

\bibitem[{Newman-Griffis and Fosler-Lussier(2019)}]{Newman-Griffis2019louhi}
Denis Newman-Griffis and Eric Fosler-Lussier. 2019.
\newblock \href {https://doi.org/10.18653/v1/D19-6218} {{Writing habits and
  telltale neighbors: analyzing clinical concept usage patterns with
  sublanguage embeddings}}.
\newblock In \emph{Proceedings of the Tenth International Workshop on Health
  Text Mining and Information Analysis (LOUHI 2019)}, pages 146--156, Hong
  Kong. Association for Computational Linguistics.

\bibitem[{Newman-Griffis et~al.(2018)Newman-Griffis, Lai, and
  Fosler-Lussier}]{Newman-Griffis2018repl4nlp}
Denis Newman-Griffis, Albert~M Lai, and Eric Fosler-Lussier. 2018.
\newblock \href {http://aclweb.org/anthology/W18-3026} {{Jointly Embedding
  Entities and Text with Distant Supervision}}.
\newblock In \emph{Proceedings of The Third Workshop on Representation Learning
  for NLP}, pages 195--206. Association for Computational Linguistics.

\bibitem[{Nguyen et~al.(2020)Nguyen, Liakata, DeDeo, Eisenstein, Mimno,
  Tromble, and Winters}]{Nguyen2020}
Dong Nguyen, Maria Liakata, Simon DeDeo, Jacob Eisenstein, David Mimno, Rebekah
  Tromble, and Jane Winters. 2020.
\newblock \href {https://doi.org/10.3389/frai.2020.00062} {{How We Do Things
  With Words: Analyzing Text as Social and Cultural Data}}.
\newblock \emph{Frontiers in Artificial Intelligence}, 3:62.

\bibitem[{{Ohio Supercomputer Center}(1987)}]{osc}
{Ohio Supercomputer Center}. 1987.
\newblock \href {http://osc.edu/ark:/19495/f5s1ph73} {Ohio supercomputer
  center}.

\bibitem[{Rahmani et~al.(2020)Rahmani, Davoudi-Monfared, Nourian, Nabiee,
  Sadeghi, Khalili, Abbasian, Ghiasvand, Seifi, Hasannezhad, Ghaderkhani,
  Mohammadi, and Yekaninejad}]{Rahmani2020}
Hamid Rahmani, Effat Davoudi-Monfared, Anahid Nourian, Morteza Nabiee, Setayesh
  Sadeghi, Hossein Khalili, Ladan Abbasian, Fereshteh Ghiasvand, Arash Seifi,
  Malihe Hasannezhad, Sara Ghaderkhani, Mostafa Mohammadi, and Mir~Saeed
  Yekaninejad. 2020.
\newblock \href {https://doi.org/10.1007/s40199-020-00369-2} {{Comparing
  outcomes of hospitalized patients with moderate and severe COVID-19 following
  treatment with hydroxychloroquine plus atazanavir/ritonavir.}}
\newblock \emph{Daru : journal of Faculty of Pharmacy, Tehran University of
  Medical Sciences}, 28(2):625--634.

\bibitem[{Runge and Hovy(2020)}]{runge-hovy-2020-exploring}
Andrew Runge and Eduard Hovy. 2020.
\newblock \href {https://doi.org/10.18653/v1/2020.blackboxnlp-1.20} {{Exploring
  Neural Entity Representations for Semantic Information}}.
\newblock In \emph{Proceedings of the Third BlackboxNLP Workshop on Analyzing
  and Interpreting Neural Networks for NLP}, pages 204--216, Online.
  Association for Computational Linguistics.

\bibitem[{Schlechtweg et~al.(2020)Schlechtweg, McGillivray, Hengchen,
  Dubossarsky, and Tahmasebi}]{schlechtweg-etal-2020-semeval}
Dominik Schlechtweg, Barbara McGillivray, Simon Hengchen, Haim Dubossarsky, and
  Nina Tahmasebi. 2020.
\newblock \href {https://www.aclweb.org/anthology/2020.semeval-1.1}
  {{{\{}S{\}}em{\{}E{\}}val-2020 Task 1: Unsupervised Lexical Semantic Change
  Detection}}.
\newblock In \emph{Proceedings of the Fourteenth Workshop on Semantic
  Evaluation}, pages 1--23, Barcelona (online). International Committee for
  Computational Linguistics.

\bibitem[{Smilkov et~al.(2016)Smilkov, Thorat, Nicholson, Reif, Viégas, and
  Wattenberg}]{embedding_projector}
Daniel Smilkov, Nikhil Thorat, Charles Nicholson, Emily Reif, Fernanda~B.
  Viégas, and Martin Wattenberg. 2016.
\newblock \href {http://arxiv.org/abs/1611.05469} {Embedding projector:
  Interactive visualization and interpretation of embeddings}.

\bibitem[{Spadea et~al.(2020)Spadea, Carraro, Saglio, Vassallo, Pessolano,
  Berger, Scolfaro, Grassitelli, and Fagioli}]{Spadea2020}
Manuela Spadea, Francesca Carraro, Francesco Saglio, Elena Vassallo, Rosanna
  Pessolano, Massimo Berger, Carlo Scolfaro, Sergio Grassitelli, and Franca
  Fagioli. 2020.
\newblock \href {https://doi.org/10.1111/tid.13470} {{Successfully treated
  severe COVID-19 and invasive aspergillosis in early hematopoietic cell
  transplantation setting.}}

\bibitem[{Talavera et~al.(2020)Talavera, Garc{\'{i}}a-Azor{\'{i}}n,
  Mart{\'{i}}nez-P{\'{i}}as, Trigo, Hern{\'{a}}ndez-P{\'{e}}rez,
  Valle-Pe{\~{n}}acoba, Sim{\'{o}}n-Campo, de~Lera, Chavarr{\'{i}}a-Miranda,
  L{\'{o}}pez-Sanz, Guti{\'{e}}rrez-S{\'{a}}nchez, Mart{\'{i}}nez-Velasco,
  Pedraza, Sierra, G{\'{o}}mez-Vicente, Guerrero, and Arenillas}]{Talavera2020}
Blanca Talavera, David Garc{\'{i}}a-Azor{\'{i}}n, Enrique
  Mart{\'{i}}nez-P{\'{i}}as, Javier Trigo, Isabel Hern{\'{a}}ndez-P{\'{e}}rez,
  Gonzalo Valle-Pe{\~{n}}acoba, Paula Sim{\'{o}}n-Campo, Mercedes de~Lera, Alba
  Chavarr{\'{i}}a-Miranda, Cristina L{\'{o}}pez-Sanz, Mar{\'{i}}a
  Guti{\'{e}}rrez-S{\'{a}}nchez, Elena Mart{\'{i}}nez-Velasco, Mar{\'{i}}a
  Pedraza, {\'{A}}lvaro Sierra, Beatriz G{\'{o}}mez-Vicente, {\'{A}}ngel
  Guerrero, and Juan~Francisco Arenillas. 2020.
\newblock \href {https://doi.org/10.1016/j.jns.2020.117163} {{Anosmia is
  associated with lower in-hospital mortality in COVID-19}}.
\newblock \emph{Journal of the Neurological Sciences}, 419.

\bibitem[{van~der Maaten and Hinton(2008)}]{tsne}
Laurens van~der Maaten and Geoffrey Hinton. 2008.
\newblock \href {http://jmlr.org/papers/v9/vandermaaten08a.html} {Visualizing
  data using t-sne}.
\newblock \emph{Journal of Machine Learning Research}, 9(86):2579--2605.

\bibitem[{Vylomova et~al.(2019)Vylomova, Murphy, and Haslam}]{Vylomova2019}
Ekaterina Vylomova, Sean Murphy, and Nicholas Haslam. 2019.
\newblock \href {https://www.aclweb.org/anthology/W19-4704} {{Evaluation of
  Semantic Change of Harm-Related Concepts in Psychology}}.
\newblock In \emph{Proceedings of the 1st International Workshop on
  Computational Approaches to Historical Language Change}, pages 29--34,
  Florence, Italy. Association for Computational Linguistics.

\bibitem[{Wang et~al.(2020{\natexlab{a}})Wang, Lo, Chandrasekhar, Reas, Yang,
  Eide, Funk, Kinney, Liu, Merrill, and Others}]{wang2020cord}
Lucy~Lu Wang, Kyle Lo, Yoganand Chandrasekhar, Russell Reas, Jiangjiang Yang,
  Darrin Eide, Kathryn Funk, Rodney Kinney, Ziyang Liu, William Merrill, and
  Others. 2020{\natexlab{a}}.
\newblock {CORD-19: The Covid-19 Open Research Dataset}.
\newblock \emph{arXiv preprint arXiv:2004.10706}.

\bibitem[{Wang et~al.(2020{\natexlab{b}})Wang, Li, Wang, Parulian, Han, Ma, Tu,
  Lin, Zhang, Liu, and Others}]{wang2020covid}
Qingyun Wang, Manling Li, Xuan Wang, Nikolaus Parulian, Guangxing Han, Jiawei
  Ma, Jingxuan Tu, Ying Lin, Haoran Zhang, Weili Liu, and Others.
  2020{\natexlab{b}}.
\newblock {Covid-19 literature knowledge graph construction and drug
  repurposing report generation}.
\newblock \emph{arXiv preprint arXiv:2007.00576}.

\bibitem[{Warmerdam et~al.(2020)Warmerdam, Kober, and
  Tatman}]{warmerdam-etal-2020-going}
Vincent Warmerdam, Thomas Kober, and Rachael Tatman. 2020.
\newblock \href {https://doi.org/10.18653/v1/2020.nlposs-1.8} {{Going Beyond
  {\{}T{\}}-{\{}SNE{\}}: Exposing whatlies in Text Embeddings}}.
\newblock In \emph{Proceedings of Second Workshop for NLP Open Source Software
  (NLP-OSS)}, pages 52--60, Online. Association for Computational Linguistics.

\bibitem[{Weiss(2014)}]{weiss-2014-muck}
Rebecca Weiss. 2014.
\newblock \href {https://doi.org/10.3115/v1/W14-3108} {{MUCK: A toolkit for
  extracting and visualizing semantic dimensions of large text collections}}.
\newblock In \emph{Proceedings of the Workshop on Interactive Language
  Learning, Visualization, and Interfaces}, pages 53--58, Baltimore, Maryland,
  USA. Association for Computational Linguistics.

\bibitem[{Wells et~al.(2020)Wells, Doores, Couvreur, Nunez, Seow, Graham,
  Acors, Kouphou, Neil, Tedder, Matos, Poulton, Lista, Dickenson, Sertkaya,
  Maguire, Scourfield, Bowyer, Hart, O'Byrne, Steel, Hemmings, Rosadas,
  McClure, Capedevilla-pujol, Wolf, Ourselin, Brown, Malim, Spector, and
  Steves}]{Wells2020}
Philippa~M Wells, Katie~J Doores, Simon Couvreur, Rocio~Martinez Nunez, Jeffrey
  Seow, Carl Graham, Sam Acors, Neophytos Kouphou, Stuart J~D Neil, Richard~S
  Tedder, Pedro~M Matos, Kate Poulton, Maria~Jose Lista, Ruth~E Dickenson,
  Helin Sertkaya, Thomas J~A Maguire, Edward~J Scourfield, Ruth C~E Bowyer,
  Deborah Hart, Aoife O'Byrne, Kathryn J~A Steel, Oliver Hemmings, Carolina
  Rosadas, Myra~O McClure, Joan Capedevilla-pujol, Jonathan Wolf, Sebastien
  Ourselin, Matthew~A Brown, Michael~H Malim, Tim Spector, and Claire~J Steves.
  2020.
\newblock \href {https://doi.org/https://doi.org/10.1016/j.jinf.2020.10.011}
  {{Estimates of the rate of infection and asymptomatic COVID-19 disease in a
  population sample from SE England}}.
\newblock \emph{Journal of Infection}, 81(6):931--936.

\bibitem[{Wendlandt et~al.(2018)Wendlandt, Kummerfeld, and
  Mihalcea}]{Wendlandt2018}
Laura Wendlandt, Jonathan~K Kummerfeld, and Rada Mihalcea. 2018.
\newblock \href {https://doi.org/10.18653/v1/N18-1190} {{Factors Influencing
  the Surprising Instability of Word Embeddings}}.
\newblock In \emph{Proceedings of the 2018 Conference of the North American
  Chapter of the Association for Computational Linguistics: Human Language
  Technologies, Volume 1 (Long Papers)}, pages 2092--2102. Association for
  Computational Linguistics.

\bibitem[{Yamada et~al.(2020)Yamada, Asai, Sakuma, Shindo, Takeda, Takefuji,
  and Matsumoto}]{yamada-etal-2020-wikipedia2vec}
Ikuya Yamada, Akari Asai, Jin Sakuma, Hiroyuki Shindo, Hideaki Takeda,
  Yoshiyasu Takefuji, and Yuji Matsumoto. 2020.
\newblock \href {https://doi.org/10.18653/v1/2020.emnlp-demos.4}
  {{{\{}W{\}}ikipedia2{\{}V{\}}ec: An Efficient Toolkit for Learning and
  Visualizing the Embeddings of Words and Entities from {\{}W{\}}ikipedia}}.
\newblock In \emph{Proceedings of the 2020 Conference on Empirical Methods in
  Natural Language Processing: System Demonstrations}, pages 23--30, Online.
  Association for Computational Linguistics.

\bibitem[{Zhao and Bethard(2020)}]{zhao-bethard-2020-berts}
Yiyun Zhao and Steven Bethard. 2020.
\newblock \href {https://doi.org/10.18653/v1/2020.acl-main.429} {{How does
  {\{}BERT{\}}{\{}'{\}}s attention change when you fine-tune? An analysis
  methodology and a case study in negation scope}}.
\newblock In \emph{Proceedings of the 58th Annual Meeting of the Association
  for Computational Linguistics}, pages 4729--4747, Online. Association for
  Computational Linguistics.

\bibitem[{Zhordaniya et~al.()Zhordaniya, Kutuzov, and
  Kuzmenko}]{zhordaniya2020vec2graph}
Tamara Zhordaniya, Andrey Kutuzov, and Elizaveta Kuzmenko.
\newblock {Vec2graph: A python library for visualizing word embeddings as
  graphs}.
\newblock Springer.

\end{thebibliography}
\bibliographystyle{acl_natbib}

\end{document}